\documentclass{article}

\usepackage{arxiv}

\usepackage[utf8]{inputenc} 
\usepackage[T1]{fontenc}    
\usepackage[hidelinks]{hyperref}       
\usepackage{url}            
\usepackage{booktabs}       
\usepackage{amsfonts}       
\usepackage{amsmath}
\usepackage{nicefrac}       
\usepackage{microtype}      
\usepackage{lipsum}
\usepackage{graphicx}
\usepackage{anyfontsize}
\graphicspath{ {./images/} }

\usepackage{natbib} 

\title{From Frege to chatGPT: Compositionality in language, cognition, and deep neural networks}

\author{
Jacob Russin\thanks{Equal primary contribution.}\\
Department of Computer Science\\
Brown University\\
\texttt{jake\_russin@brown.edu}\\
\And
Sam Whitman McGrath\footnotemark[1]\\
Department of Philosophy\\
Fordham University\\
\texttt{smcgrath20@fordham.edu} \\
\AND
Danielle J. Williams\\
Department of Philosophy\\
Psychological and Brain Sciences\\
Washington University in St. Louis\\
}

\begin{document}
\maketitle

\makeatletter
\let\@fnsymbol\@arabic
\makeatother

\setcounter{footnote}{0}

\section{Introduction}
\label{sec:intro}
Deep neural networks (DNNs) have made breakthrough after breakthrough in artificial intelligence (AI) over the last decade, reproducing sophisticated cognitive behaviors from advanced gameplay in board games like Go \citep{SilverHuangMaddisonEtAl16}, Chess \citep{SilverHubertSchrittwieserEtAl18}, and Diplomacy \citep{METAFUNDAMENTALAIRESEARCHDIPLOMACYTEAMFAIRBakhtinBrownEtAl22}, to catalyzing achievements in mathematics \citep{Romera-ParedesBarekatainNovikovEtAl24}, science \citep{JumperEvansPritzelEtAl21}, engineering \citep{MerchantBatznerSchoenholzEtAl23}, and medicine \citep{OmiyeGuiRezaeiEtAl23, SinghalAziziTuEtAl23}. 
The large language models (LLMs) powering AI products like chatGPT \citep{BrownMannRyderEtAl20a, BubeckChandrasekaranEldanEtAl23, OpenAIAchiamAdlerEtAl24} have shown remarkable proficiency in cognitive domains previously regarded as uniquely human, such as natural language syntax \citep{LinzenBaroni21a}, critical reasoning and argumentation \citep{HerboldHautli-JaniszHeuerEtAl23}, and computer programming \citep{ChenTworekJunEtAl21, BubeckChandrasekaranEldanEtAl23}. 

In humans, these creative cognitive behaviors have traditionally been explained by postulating a special cognitive property: compositionality.
The precise definition of compositionality varies by discipline, but intuitively it can be understood as the ability to compose familiar constituents (words, concepts, percepts) into novel complex combinations.
In particular, cognitive scientists have often modeled human generalization behaviors with classical architectures that intrinsically guarantee that familiar constituents can be redeployed in novel constructions \citep{FodorPylyshyn88}.
Theorists have argued that this explanatory framework is ‘non-negotiable,' that compositionality is the ``only game in town'' for explaining the power and flexibility of human intelligence across many domains, including natural language syntax and semantics \citep{Fodor75}.

Unlike classical symbolic AI systems, however, the neural networks powering  recent advances are not designed to embody a strong assumption of compositionality, relying instead on acquiring capabilities by learning statistical regularities from data \citep{FodorPylyshyn88,  McClellandRumelhartthePDPResearchGroup86, RumelhartMcClellandPDPResearchGroup86, SmolenskyMcCoyFernandezEtAl22c}. 
Whereas compositionality is integral to certain fundamental design features of classical computational architectures (e.g., predicate-argument structure, variable binding, constituent structure), these features are not native to standard DNNs. 
Traditionally, this has been lodged as an objection to the viability of neural networks as cognitive models \citep{FodorPylyshyn88}. 
However, modern DNNs’ impressive performance in domains traditionally thought to require compositionality (e.g., language) calls this into question. 
Have today’s models overcome previously hypothesized limitations? 
Or are they merely imitating compositional behaviors through useful but imperfect heuristics? 
Do DNNs instantiate a novel form of non-classical compositionality?
Or does their success in reproducing sophisticated cognitive behaviors show that there is a \emph{new} game in town and that compositionality \emph{is} negotiable after all?   

In this chapter, we’ll review and contextualize recent studies of compositionality in DNNs, with a particular focus on language. 
We begin in section \ref{sec:frege_to_fodor} with a historical overview of the explanatory role of compositionality in philosophy, linguistics, and psychology, outlining its evolution from a property of interest primarily to logicians and philosophers to an empirically motivated psychological construct central to contemporary cognitive science. 
We then briefly describe the rise of connectionism in the 1980s and 90s (section \ref{sec:connectionism}) and the challenge that compositionality poses for neural networks (section \ref{sec:FandP}). 
With this crucial context in place, we move to today’s deep neural networks, reviewing key recent breakthroughs in replicating compositional behaviors (section \ref{sec:review}).
In particular, we argue that metalearning (section \ref{sec:metalearning}), or learning to learn, offers a novel perspective on how neural networks like LLMs (section \ref{sec:llms}) can come to reproduce the behavioral signatures of compositionality. 
To close, we discuss potential implications that this work may have for our understanding of human compositionality, and suggest avenues for future empirical and philosophical work (section \ref{sec:discussion}). 

\section{Compositionality in Context}
\label{sec:comp_in_context}

\subsection{From Frege to Fodor}
\label{sec:frege_to_fodor}
Though the compositionality of human language and thought is a central commitment of contemporary cognitive science, its current explanatory role was not always evident. 
Appreciation of the significance of compositionality first emerged at a time when philosophy, psychology, logic, and linguistics were still intertwined, and it took nearly a century of philosophical investigation to cleanly excavate the question(s) that the compositionality principle purports to answer.\footnote{Compositionality can be traced even further back in the history of philosophy, to Aristotle and the Sanskrit grammarians \citep{PaginWesterstahl10a}, but our focus here will be on the 20th century analytic tradition, in which this phenomenon was first named and analyzed  \citep{Putnam95, KatzFodor63}.}

Gottlob Frege made indispensable contributions to the disentangling of these domains and was arguably the first philosopher to fully appreciate the explanatory potential of the compositionality principle \citep{Dummett93}. 
In contemporary terms, the compositionality principle states that the meaning of a complex expression is determined by the meaning of its parts and the way they are combined. 
There are a number of different ways of precisifying this principle,\footnote{For discussion, see \citet{PaginWesterstahl10a} and \citet{Partee04}.}
and two importantly different ways of interpreting its significance---as a claim about the metaphysics of meaning or as a claim about the psychological processes with which humans generate and grasp meanings. 
On the first interpretation, the compositionality principle asserts that the meaning of a complex expression (e.g. a sentence) is determined by or \emph{supervenes on} the meanings of its parts (e.g. words), which are explanatorily or metaphysically fundamental.
On the second, it expresses a psychological hypothesis about how language users themselves determine this meaning---namely, by assembling the meanings of the parts in accordance with tacitly known rules \citep{Dummett96}.\footnote{These two readings might be tethered to a distinction between two different senses of “determination.” Consider an analogy. Objectively speaking, the solution to the equation ``a x b = ?'' is determined by the values of ``a'' and ``b'' (as well as the nature of multiplication). Psychologically speaking, however, there are numerous ways that someone might determine this solution---by searching their memory, by guessing, or by actually multiplying the component values. Our question is which of these senses of determination, the objective or the subjective sense (so to speak), is involved in the compositionality principle.}

Though Frege clearly appreciated the significance of compositionality and anticipated key contemporary arguments in its favor \citep{Frege56, Frege93}, it is not clear which of these two construals of the compositionality principle he intended. 
In fact, though it has been dubbed “Frege’s Principle” \citep{Cresswell73}, Frege never explicitly formulated the compositionality principle himself. The first philosopher to do so was Frege’s one-time student, Rudolf Carnap, who named it after his former teacher \citep{Carnap88}.

Carnap was the leading proponent of logical empiricism and, among other contributions, paved the way for what was arguably the central research program in 20th century formal semantics \citep{Carnap88, Montague74}. 
Carnap employed ``Frege's Principle'' as a crucial theoretical assumption. 
However, like Frege before him, Carnap was deeply skeptical that compositional formal languages could be treated as models of natural languages, like German, English, or Mandarin \citep{Carnap37}. 
His theoretical ambition was not to explain everyday language use or uncover underlying psychological mechanisms, as a contemporary theorist might expect, but to develop artificial languages ideally suited to mathematical and scientific purposes \citep{Harris17}.
Compositionality was an indispensable feature of these artificial languages, but Carnap was well aware of the idealizations that they involved---abstracting away from central features of natural language, like context-sensitivity and ambiguity---and he consciously avoided putting them forward as models of ``vernacular'' human language.
This was true for philosophers in the ‘ideal language tradition’ more broadly. 
They appreciated the power of mathematical logic and the importance of the compositionality principle, but did not advance their formal theories as psychological models or treat compositionality as an empirical hypothesis about human linguistic processing.

Ordinary Language Philosophy, the other major offshoot of the analytic tradition's ‘linguistic turn’ \citep{Rorty92}, criticized the ideal language philosophers for this abstraction and focused their analyses on the messy realities of natural language use \citep{Ryle49}. 
Leaving aside the movement’s grander ambitions \citep{zotero-63660, Horwich13}, philosophers in the ordinary language tradition made clear-cut contributions to linguistic theorizing, from J.L. Austin’s work on speech acts \citep{Austin75} to H.P. Grice’s pioneering contributions to the field of pragmatics \citep{Grice91}. 
However, these philosophers were either skeptical or silent about the compositionality of natural language \citep[see, for example, Gilbert Ryle's attack on the ‘building block' theory of meaning in][]{Ryle57}. 

The tension between these approaches persisted into the early 1960s, with ideal language philosophers developing compositional  models, but abstracting away from natural language use, and ordinary language philosophers advancing analyses of natural language, but rejecting or simply ignoring the principle of compositionality. 
This contrast formed the philosophical backdrop for Noam Chomsky’s revolution in the field of syntax \citep{Chomsky57, Chomsky65}, which would help inaugurate the ‘cognitive turn' and fundamentally alter the role of compositionality in linguistic theorizing.

Chomsky rejected the key assumption that both groups of philosophers shared---that formal mathematical languages could not be treated as psychological models of ordinary linguistic knowledge. 
This allowed him to synthesize their approaches, using the “methodological toolkit” developed by philosophers interested in formal logic, from Frege onwards \citep{Harris17}, but taking them seriously as models of the syntax of natural languages \citep{Chomsky57, Katz71}. 
Chomsky did not focus on questions of meaning, but the synthesis that his work enacted was a crucial step in the emergence of the compositionality principle as an empirical hypothesis about human linguistic processing.
Indeed, Chomsky's synthesis was promptly extended to formal semantics by Richard Montague and other key contributors \citep{KatzFodor63, Davidson65}. 
Montague explicitly argued that there is “no important theoretical difference between natural languages and the artificial languages of logicians,” allowing him to treat rule-based semantic systems of the kind pioneered by Carnap as psychological models of human linguistic competence \citep[p. 15]{Montague74}.\footnote{Montague signals his alignment with Chomsky on this key point: ``I consider it possible to comprehend the syntax and semantics of both kinds of languages within a single natural and mathematically precise theory. On this point I differ from a number of philosophers, but agree, I believe, with Chomsky and his associates'' \citep[p.15]{Montague74}.}

It would be hard to overstate the impact that this shift had on linguistic theorizing. 
In effect, it transformed formal syntactic and semantic theories from abstract objects of study for logicians, mathematicians, and philosophers into psychological constructs that underlie and explain the linguistic capacities of even the youngest children capable of speech. 
With this shift, the compositionality principle took on a newfound significance and key role in explaining human linguistic behavior.

Chomsky drew attention to the creativity of human language use and its open-ended, potentially infinite expressive capacity, which must be achieved through finite means \citep{Chomsky66, Chomsky20, Frege93}. 
To use a classic example discussed by \citet{KatzFodor63}, in even the most routine linguistic interactions we regularly encounter novel sentences, yet fluidly and effortlessly comprehend their meaning and generate appropriate responses. 
This open-ended creative capacity seems to cry out for explanation.\footnote{Chomsky suggests that some 17th century philosophers, like Descartes and Wilhelm von Humboldt, recognized this explanatory challenge and anticipated his own ideas \citep{Chomsky66}. It is also prefigured by Frege, who wrote in 1914 that, “the possibility of our understanding sentences which we have never heard before rests evidently on this, that we can construct the sense of a sentence out of parts that correspond to words… Without this, language in the proper sense would be impossible… we would always be restricted to a very narrow area and could not form a completely new proposition” \citep{Frege93}.}
How are humans capable of understanding sentences that they have never before encountered? 

The answer, in one word, is compositionality. 
If the meaning of an unfamiliar sentence is determined by the meaning of its parts, their mode of combination, and nothing else, then knowing the meanings of component words and the grammar of the language will suffice for grasping this novel meaning. 
Appealing to tacit knowledge \citep{Fodor68a, Evans96} thus provides an explanation of the human capacity to understand novel sentences---we derive the meaning of novel constructions by recombining known components according to known syntactic rules.

In fact, it can seem that tacit knowledge of a compositional grammar is the \emph{only} adequate explanation of this capability. 
Chomsky argued that humans’ creative capacity is essentially unbounded \citep{Chomsky57}. 
To take a classic example, if we idealize away from constraints of memory and attention (as well as the finitude of a human lifetime!), a compositional subject could go on adding new conjuncts to the end of a well-understood sentence \emph{ad infinitum}. 
Provided that they understand the meaning of the new conjunct (and conjunction itself), they will also understand these newly formed sentences. 
It is this open-endedness that makes an internalized compositional system seem like ``the only game in town.''
Alternative explanatory approaches of the time, like behaviorism, struggled to account for this form of unbounded extension.
For Chomsky, Montague, Fodor, and others, the solution was clear: posit internal compositional mechanisms that support these behavioral capacities. 

This abductive inference to the best explanation forges a close link between creativity, productivity, and compositionality.
Particular observations about productive or systematic behaviors in humans motivate the ascription of internal compositional mechanisms to ordinary speakers.
Such arguments commit theorists to interpreting the compositionality principle as a psychological, rather than a metaphysical, thesis.\footnote{To return to our analogy, if we aim to explain the behavior of someone solving a multiplication problem, what matters is how \emph{they} determined the solution, in the subjective sense, rather than the metaphysical or objective grounds of this solution.} 
This point deserves emphasis. 
To serve as a robust explanation of human language comprehension and observable linguistic behavior, these internal compositional mechanisms must be `psychologically real' and actually employed as humans determine the meanings of novel expressions.\footnote{This observation may seem obvious, but it is in fact easy to overlook. Even Davidson, one of the earliest and most influential proponents of compositional natural language semantics, fails to fully appreciate the point. For a detailed discussion of this tension in Davidson's work, see \citet{zotero-63695}.}
For the proposed explanation to work, it is not enough that the language itself admits of a compositional analysis---after all, one and the same language may admit both compositional and non-compositional models.

To see the point, consider a simplified `toy' language \citep{Evans96} with only 10 names (a, b, c...) and 10 predicates (F, G, H...), generating 100 possible sentences (Fa, Fb, Ga...). This language can be provided with a compositional model, along familiar lines. However, every admissible sentence in this language could also be generated by a very different, though `extensionally equivalent,' grammar that has 100 different axioms specifying the meaning of each. How can we tell which of these models a speaker of this minimal language possesses? Many philosophers, from W. V. O. \citet{Quine70} to Crispin \citet{Wright81}, were skeptical that such questions could be made empirically or behaviorally tractable. 
\citet{Evans96}, however, argues that there are clear empirical considerations that justify the ascription of the compositional system rather than the non-compositional one. 
In principle, we could scrutinize the speaker's internal states and attempt to detect compositional structure. 
In practice, (and in keeping with the methodological assumptions of researchers in the Chomskyan tradition) we can instead rely on a wider class of behavior---compositional generalization behavior involving the introduction of novel names or predicate. 
If you introduce a new sentence, Fz, will the speaker be able to understand Gz and Hz?
If so, this is evidence that they understand the language compositionally.
Evans' proposal about how to operationalize compositionality is very similar to those used in contemporary machine learning research (see section \ref{sec:operationalizing}), and it demonstrates the central role that linguistic \emph{behavior} plays in motivating \emph{cognitive} explanation in terms of the compositionality principle.

These theoretical developments close the gap between early philosophical theorizing about compositionality and the ongoing debates in contemporary cognitive science. 
Compositionality is understood as an explanatory psychological property, operationalized in terms of the specific kinds of productive and systematic generalization behaviors that humans exhibit \citep{Evans96}.
In each case, the characteristic `behavioral signatures' of compositionality are taken as evidence diagnostic of internalized compositional mechanisms, which serve to explain the behaviors in question. 
As Chomsky often emphasizes, this means that even the most abstract reaches of formal linguistics are properly understood as a branch of cognitive psychology \citep[and ultimately biology, though developing an account of how symbolic systems are implemented in the brain is sometimes treated as a subsidiary enterprise; see][]{Chomsky68}.

Two further consequences of the Chomskyan revolution deserve mention before concluding this historical overview. 
First, the rise of this new explanatory paradigm brought with it a notable resurgence of nativism, as Chomsky and others argued that the linguistic data to which children are exposed is insufficient for recovering the kind of fine-grained syntactic structures necessary for explaining human linguistic competences \citep{LaurenceMargolis01}. 
As we will briefly discuss in \ref{sec:development}, the rise of DNNs and LLMs may make novel contributions to this long-standing philosophical issue \citep{Buckner18, Buckner19, Buckner23a}.
And second, though we have framed the discussion here in terms of the meaning of sentences, this period saw a distinct shift in focus from language to thought. 
This `cognitive turn’ transposed key debates about language into questions about the structure of cognition itself \citep{Burge07}. 
Compositionality was no exception. 
Jerry Fodor influentially argued that the compositional structure of natural languages is a reflection of the compositional structure of thought, reintroducing the language of thought hypothesis to the philosophical mainstream \citep{Fodor75, PelletierRoques17}. 
On this view, human cognition is compositional from the ground up---it is a feature of the mind’s basic operating system, not just an artifact of one specific domain (language). 

Across this historical evolution, compositionality arrives in the position familiar to cognitive scientists today: a core property of human cognition, motivated by its non-negotiable role in explaining the creativity, systematicity and productivity of language and thought. 
From this perspective, the inability to adequately account for compositionality is a fatal flaw, which Fodor wielded like a sledgehammer to smash rival theories of meaning and defend his language of thought hypothesis. 
Over the years, he would take this sledgehammer to the use theory of meaning \citep{FodorLepore01, Fodor03}, conceptual role semantics \citep{FodorLePore93}, prototype theory \citep{FodorLepore96}, and most importantly for our present purposes, neural network models of cognition \citep{FodorPylyshyn88}.

\subsection{The Rise of Connectionism}
\label{sec:connectionism}
In the 80s and 90s, a series of key advances in neural networks, such as the invention of Hopfield networks \citep{Hopfield82} and the backpropagation algorithm \citep{RumelhartHintonWilliams86}, led many to abandon the classical, Fodorian picture and turn to an alternative view of the mind \citep{McClellandRumelhartthePDPResearchGroup86, RumelhartMcClellandPDPResearchGroup86}. 
The connectionist movement was initially motivated by considerations other than compositionality, such as the basic computational structure of biological brains. 
Proponents sought to demonstrate that sophisticated cognitive behaviors could emerge from a network of simple interconnected units (or “neurons”). 
Though these units abstract away from a great deal of what is known about biological neurons, they implement a similar kind of computation: each unit performs a weighted sum of its inputs and applies a threshold or nonlinearity, much as biological neurons integrate dendritic inputs and spike at a certain threshold \citep{HodgkinHuxley52, McCullochPitts43, OReillyMunakataFrankEtAl12}. 
Early work demonstrated how these networks could learn by incrementally updating their weights (or ``synapses'') to improve overall performance on a given task \citep{RumelhartMcClellandPDPResearchGroup86}.

Connectionists sought to construct models that explained behavioral phenomena without building in classical representational elements (e.g., rules, variables, symbols, propositions, constituent structure), instead emphasizing how useful representations could emerge in a neural networks’ distributed patterns of activity over the course of learning \citep{RumelhartMcClelland86a}. 
This approach stood in tension with existing explanations based on the principle of compositionality, since standard distributed representations have no natural notion of “parts” or “constituents” \citep[although key later work by][would explore this possibility]{Smolensky90a}. 
The network’s objective is to discover and deploy whatever representational elements are most useful for solving a given problem. 
These emergent representations may not align with the kinds of features a linguistic or cognitive theorist might put forward, and may not even afford succinct interpretations in psychological terms at all \citep{Smolensky86}. 
In fact, as we have discussed elsewhere \citep{McGrathRussinPavlickEtAl23}, many early connectionists were eliminativists about classical symbolic constructs \citep{RumelhartMcClelland86a, RamseyStichGaron90, McClellandPatterson02a}, arguing that the rise of neural networks repudiated traditional approaches to cognitive science and constituted a paradigm shift for the field \citep{McClellandRumelhartHinton86}.

Neural networks thus seemed to offer a radically new approach to the study of cognition, providing novel explanations for many different kinds of behavioral phenomena \citep{RumelhartMcClellandPDPResearchGroup86}, including perceptual processes \citep{McClellandRumelhart81}, memory processes \citep{Hopfield82, McClellandMcNaughtonOReilly95}, executive and decision-making processes \citep{CohenDunbarMcClelland90, MillerCohen01}, and language processes \citep{Dell86, RumelhartMcClelland86a}. 
The emergence of the connectionist movement generated a great deal of excitement and controversy across the cognitive sciences, including within philosophy \citep{RamseyRumelhartStitch}, where Fodor was waiting (with his sledgehammer).

\subsection{The Compositionality Challenge}
\label{sec:FandP}
In response to the threat posed by connectionism, Fodor penned a long critique of the movement with the psychologist, Zenon Pylyshyn \citep{FodorPylyshyn88}. 
Their highly influential article, ``Connectionism and Cognitive Architecture: A Critical Analysis,'' develops a dilemma for proponents of neural network models of cognition.\footnote{Framing Fodor and Pylyshyn's argument as a dilemma is common in the literature \citep{MilliereBuckner24} and is consonant with the earliest connectionist replies to their article \citep{Smolensky91}, but one could argue that the two horns of this dilemma really comprise a single objection---that neural networks cannot explain compositionality. Our goal here is not to engage in `Fodorology.' Dividing what we will call the `compositionality challenge' into two distinct horns is, in our view, a useful way of distinguishing two importantly different objections to DNNs, which often get misleadingly run together.} 
The first horn of this dilemma is the claim that core properties of neural networks make them incapable of replicating the behavioral signatures of compositionality, rendering them empirically inadequate as models of human cognition.
This claim has real predictive upshot---if it is correct, neural networks should fail to match human performance on the specific tasks that demand compositional generalization \citep{Marcus98}.
The second horn, which we will call the `mere implementation' objection, is subtler and involves more overtly philosophical considerations.
It alleges that, even if DNNs were to replicate the behavioral signatures of compositionality, this would only mean that they had \emph{merely implemented} a classical symbolic architecture \citep{McGrathRussin2024}, rather than providing a genuine alternative explanation.
Taken together, these two horns comprise the compositionality challenge for artificial neural networks.

This first horn of Fodor and Pylyshyn’s argument draws on two key cognitive properties that are taken to require compositionality: productivity and systematicity. 
As we have seen, classical symbolic accounts provide a natural mechanism for productively generating an unbounded set of novel thoughts through the recombination of a finite pool of primitive elements. 
Connectionist architectures, on the other hand, do not seem to build in the kind of combinatorial constituent structure capable of supporting this form of unbounded productivity. 
They also struggle to capture the systematicity of cognition---the apparently intrinsic connection between the ability to comprehend and generate specific sets of thoughts/sentences. 
To use the classic example, it seems that anyone capable of understanding the sentence ``John loves Mary'' should be capable of understanding the sentence ``Mary loves John.'' 
The classical account again provides a natural explanation---the meaning of the second sentence is a function of the very same components and mode of combination as the first. 
Therefore, any compositional learner capable of understanding the first will be able to redeploy this knowledge in understanding the second.
Fodor and Pylyshyn argue that in a connectionist network without constituent structure, on the other hand, each of these individual acts of understanding would be an independent capacity. 
There would therefore be no inherent reason why they should hang together---a network trained on one could easily misunderstand the other. 
For both productivity and systematicity, Fodor and Pylyshyn’s conclusions are the same---without underlying compositional mechanisms, connectionist networks will fail to replicate these core features of human language and cognition.

The second horn of Fodor and Pylyshyn's dilemma draws on the possibility of implementing a classical symbolic system in an artificial neural network. 
Even if a connectionist network did manage to exhibit systematic and productive behaviors, they argue, this would have to be because it had merely implemented a classical symbolic system, undermining any claim to revolutionary implications or novel insight into the nature of the mind (for further discussion, see section \ref{sec:FandP2}). 
Thus, the connectionist is left with no way out: either their models are empirically inadequate, or they are mere implementations, which may prove informative for theorists interested in how symbolic systems are realized in the human brain \citep{McLaughlin93}, but will not contribute to higher-level cognitive theory. 
As should be clear, this argument is a species of the genus described in section \ref{sec:frege_to_fodor}---compositionality is treated as an empirically motivated psychological construct, necessary for the explanation of certain key human behaviors. 

The compositionality challenge ignited a heated debate and drew a wide array of rebuttals and clarifications \citep{Chalmers93, Hadley97, HorganTienson91, Matthews94, RamseyStichGaron90}. 
Some connectionist researchers---most notably Paul Smolensky---tried to meet the `mere implementation’ objection head on by building neural networks with internal compositional structure that did not merely recapitulate classical architectures \citep{Smolensky90a, Smolensky91}. 
This led to further debate about just what kind of explanation compositionality offered and what internal compositional structure was rightly considered ‘non-negotiable’ \citep{MacDonaldMacDonald91, vanGelder90}.
Our primary goal in this paper---and the sole focus of the review in section \ref{sec:review}---is to survey recent findings suggesting that contemporary DNNs have plausibly surmounted the first horn of this dilemma.
In \ref{sec:FandP2}, we also briefly evaluate whether this means that they fall on the second horn \citep[for further discussion, see][]{McGrathRussinPavlickEtAl23, McGrathRussin2024}.

Over time, the debates over connectionism waned, moving from the center to the periphery of cognitive science. 
Neural networks, however, continued to progress, and exploded into new prominence with the inception of the deep learning revolution \citep{KrizhevskySutskeverHinton12, LeCunBengioHinton15, GoodfellowBengioCourville16}, instilling the debates over connectionism and compositionality with renewed importance.

\section{In-Context Compositionality}
\label{sec:review}

Many advances in neural networks have accompanied the deep learning revolution, including the introduction of the transformer architecture \citep{VaswaniShazeerParmarEtAl17a}.
However, the single most important factor that differentiates modern neural networks from their connectionist predecessors is their scale: modern networks are deeper (i.e. more layers), have many orders of magnitude more learnable parameters (weights), and are trained on many orders of magnitude more data. 
This is possible due to a combination of advances in hardware (e.g., GPUs) and the availability of large amounts of data from the internet (e.g., Wikipedia). 
These are merely quantitative differences, but they have led to qualitative increases in these systems' capabilities \citep{BrownMannRyderEtAl20a, CaballeroGuptaRishEtAl23, KaplanMcCandlishHenighanEtAl20, WeiTayBommasaniEtAl22}.

Although these developments have led to remarkable gains in performance across many domains, it is unclear \emph{a priori} whether they should have a significant impact on DNNs' underlying compositionality. 
The features of neural networks that led theorists like Fodor and Pylyshyn to conclude that they were not compositional seem \emph{prima facie} to be largely unchanged in modern deep networks: DNNs are still extremely unbiased learners with no explicit rule-like representations or built-in sensitivity to constituent structure.\footnote{While most agree that the transformer is a very unbiased architecture \cite[e.g., it is in some ways even more unbiased than the convolutional neural network;][]{DosovitskiyBeyerKolesnikovEtAl21}, some have argued that the attention mechanism might play a special role in the compositional behaviors observed in transformers \citep{SmolenskyMcCoyFernandezEtAl22c}.}
As we have seen, accounting for compositionality is a pressing theoretical challenge---if neural networks are to provide illuminating accounts of human cognitive capacities, they must be able to model the creative, productive, and systematic generalization behaviors exhibited by humans. 

Our central question in this review will be an empirical one: can contemporary neural networks replicate the behavioral signatures of compositionality?
We will argue that although many current neural networks (including ones leveraging modern architectures such as the transformer) still seem incapable of capturing these behavioral signatures, novel approaches have shown that it is possible to endow neural networks with the properties that are required to produce them. 
In particular, recent work has shown that metalearning neural networks, which learn how to learn new tasks given \textit{in context}, can reproduce key compositional generalization behaviors. 
These results cast doubt on the empirical predictions made by classical theorists that neural networks would never be capable of generalizing systematically or productively, and suggest that DNNs may meet the challenge posed in the first horn of Fodor and Pylyshyn's dilemma.

\subsection{Operationalizing Compositionality}
\label{sec:operationalizing}

An empirical investigation of compositionality in neural networks requires that we operationalize it by specifying exactly what kinds of behaviors it is supposed to afford---we need a set of behavioral measures that allow us to determine whether a given learner is compositional \citep{Garson94}. 
Broadly speaking, compositionality has been evaluated in DNNs by measuring the degree to which they generalize under specific conditions. 
This reflects the explanatory role played by compositionality for classical theorists \citep{KatzFodor63, Evans96}: compositionality is a property of the underlying cognitive system whose existence we can infer precisely because it explains certain kinds of generalizations (e.g., systematic and productive ones) that we observe in the system’s behavior.

\begin{figure}[t]
    \centering
    \includegraphics[width=0.5\textwidth]{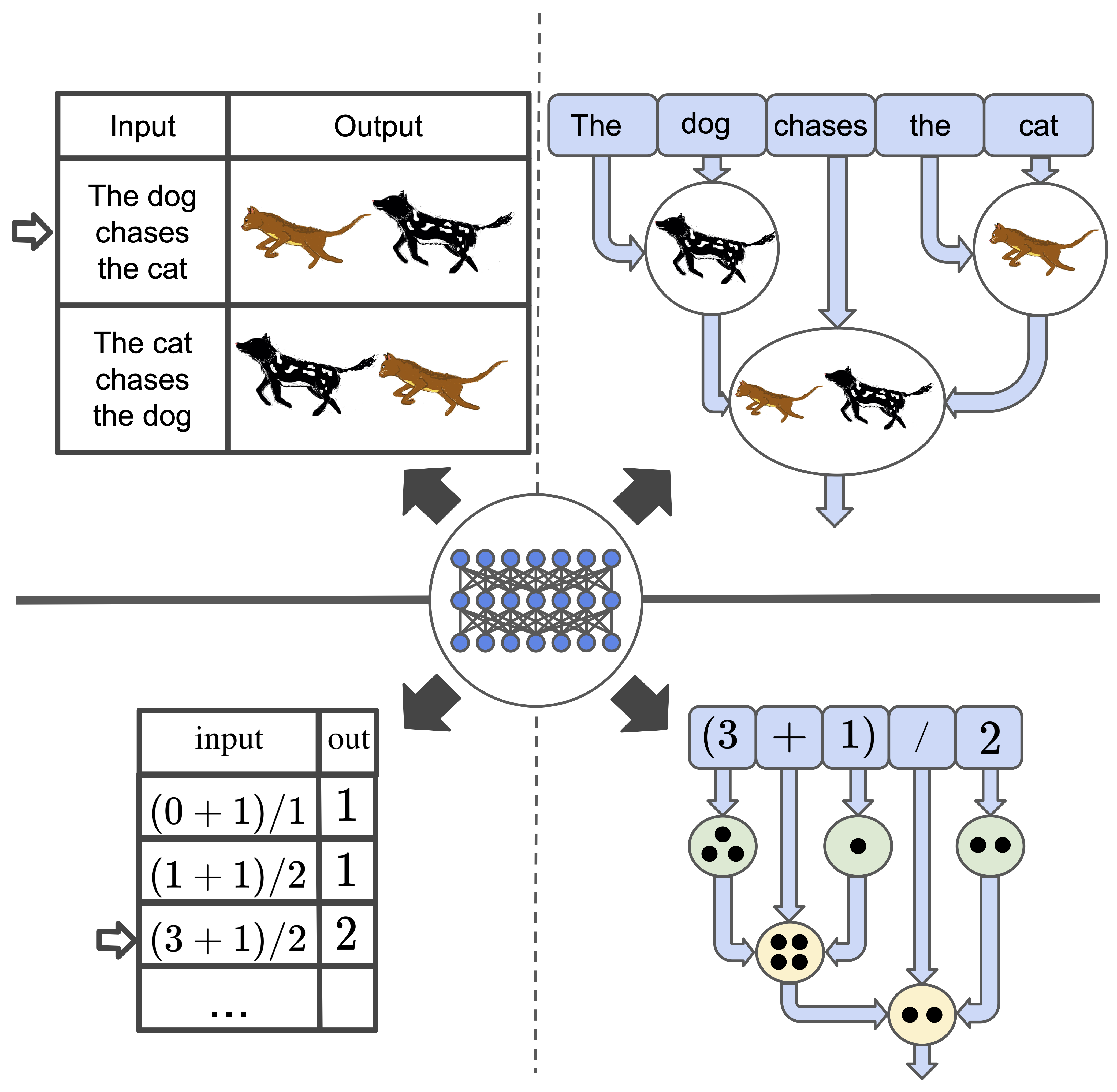}
    \caption{Are neural networks stimulus-response learners, or do they learn to solve problems with compositional algorithms? Two extreme possibilities are depicted for natural language (top) and simple arithmetic (bottom). In either case, a stimulus-response learner (left) would simply memorize a lookup table where entries corresponded to entire inputs/outputs, ignoring any compositional structure. A compositional learner (right) would solve the same problems by composing familiar elements according to known rules, allowing generalization. Figure reproduced from \citet{RussinFernandezPalangiEtAl21}.}
    \label{fig:compositionality}
\end{figure}

The behavioral tasks used to evaluate the compositionality of DNNs typically operationalize it as the ability to generalize from a given set of training samples to a given set of test samples. 
These train and test sets are crafted such that a compositional learner ought to be able to infer the underlying grammatical structures or rules that govern the training examples, and then leverage that knowledge to systematically generalize to the test samples.\footnote{An alternative way to operationalize compositionality is to test sample efficiency, or the number of training samples it takes to achieve a certain level of test performance \citep{BahdanauMurtyNoukhovitchEtAl18}. This makes a similar assumption that a compositional learner should be able to infer the rules governing the data from relatively few samples.}
For example, the model might be trained on the meanings of instructions using novel verbs, like “dax,” “flug,” and “flug twice,” and then tested on an unseen instruction “dax twice.” 
If the learner is compositional, it should be able to extract the meanings of “dax,” “flug,” and “twice” from the training samples, as well as knowledge of how adverbs modify verbs in the underlying grammar.\footnote{It would probably require more than the three examples given above to truly infer their underlying structure, but we have chosen to keep it brief for illustrative purposes.}
If the model infers this underlying structure correctly, it should be able to generalize to the test sample (“dax twice”) by composing the familiar meaning of the verb “dax” with the familiar meaning of the adverb “twice” according to the known grammatical rules \citep{LakeBaroni18}. 
On the other hand, a non-compositional learner---for example, a stimulus-response learner that simply stored each training sample in a lookup table (see Figure \ref{fig:compositionality})---would have no basis upon which to generalize to the novel expression “dax twice.”

In a seminal study on compositionality in modern DNNs, \citet{LakeBaroni18} leveraged these principles to develop the SCAN task, a sequence-to-sequence task where the model takes instructions as inputs (e.g., “walk left”), and must return the corresponding sequence of actions (e.g., TURN\_LEFT WALK). 
To test compositional generalization, the researchers created a systematic difference between the train and test sets: one of the four primitive verbs (“jump”) was left out of the training set except in its simplest form (“jump” → JUMP).
This created a scenario like the one discussed above, where a model could in principle infer the underlying grammar from all of the complicated constructions containing the other primitive verbs (e.g., “walk twice,” “look twice,” etc.), and combine this with knowledge of the action associated with “jump” (JUMP) to generalize to the analogous constructions containing “jump” (e.g., “jump twice”).

The results showed that even though standard recurrent neural networks (RNNs) performed nearly perfectly when the train and test sets were randomly split, these same networks performed quite poorly on this compositional generalization test (about 1.2\% accuracy reported in the original study).
This was true for all of the models they tested, and follow-up work showed that other standard architectures such as convolutional neural networks \citep{DessiBaroni19} and transformers \citep{KeysersScharliScalesEtAl20} exhibit similar reductions in performance on the compositional test (though less extreme in the case of the convolutional neural network). 
Analogous experiments in humans using a modified version of the same task have shown that humans perform quite well \citep{LakeLinzenBaroni19a, LakeBaroni23}. 
A number of subsequent machine learning studies followed up on this work, creating similar tests of compositional generalization and showing that these standard DNNs fail to generalize compositionally in domains such as semantic parsing \citep{KimLinzen20a}, string manipulation with a probabilistic context-free grammar \citep{HupkesDankersMulEtAl20}, mathematics \citep{SaxtonGrefenstetteHillEtAl19b}, natural-language translation \citep{DankersBruniHupkes22}, and grounded language understanding \citep{RuisAndreasBaroniEtAl20a}, among others. 

Taken together, the studies that have evaluated the compositionality of standard DNN architectures show that while they excel at generalizing when train and test data are split randomly, they perform remarkably worse when there is a systematic difference between train and test. 
One way of understanding these results is in terms of classical cognitive theory---as vindicating theorists who held that compositionality should be a necessary consequence of cognitive architecture and consequently predicted that neural networks will inevitably fail to replicate the systematicity and productivity apparent in human performance \citep{FodorPylyshyn88, Marcus98}. 

Another (perhaps compatible) way of understanding these results is to take a statistical learning theory perspective.
Neural networks are universal function approximators \citep{Cybenkot, Hornik} that assume that the train data and test data are sampled from the same distribution, i.e., that the train and test data will be independent and identically distributed (i.i.d.). 
Viewed through this lens, the generalization behaviors that are taken to be diagnostic of compositionality \citep{Evans96, FodorPylyshyn88} are out-of-domain (o.o.d.) generalizations, and require extrapolation, rather than interpolation \citep{Marcus98, Marcus18}.
For an unbiased learner like a standard neural network, this kind of generalization should be unlikely or impossible. 

The extent to which a learner is expected to generalize beyond its training distribution is the extent to which it must make additional assumptions about the world. 
These assumptions, called `inductive biases' in statistical learning parlance, can take many forms in neural networks, including architectural inductive biases or extra regularization terms in the loss function (e.g., weight decay, a regularization term that encourages a network’s weights to be small).
Standard neural networks do not make very many of these assumptions and thus tend to struggle when even the most superficial distributional shifts occur between train and test \citep{LakeUllmanTenenbaumEtAl17a}. 
For example, standard neural networks used in deep reinforcement learning agents trained on Atari games fail to generalize when superficial features of the game are changed \citep{KanskySilverMelyEtAl17}. 
It is not surprising, then, that they also fail in compositional generalization settings where there can be extreme distributional shifts from train to test. 

However, recent work has investigated methods that endow neural networks with inductive biases that would facilitate their adaptation to these shifts and allow them to replicate the behavioral signatures of compositionality. 
In the following, we’ll review recent work suggesting that neural networks can acquire such inductive biases via metalearning, or learning to learn. 
In the metalearning setting \citep{LakeBaroni23}, an inductive bias is imparted to a neural network by training it on a specific distribution of tasks (e.g., compositional generalization tasks), endowing it with an ability to learn new tasks in specific ways.
Similarly, when neural networks such as LLMs are pretrained on a large amount of data, they can acquire an emergent ability to learn new tasks given in context \citep{BrownMannRyderEtAl20a, BubeckChandrasekaranEldanEtAl23}.

In the following, we will review key findings from recent studies exploring compositionality in each of these settings, highlighting how they challenge the traditional view of compositionality in neural networks. 
Our review will largely focus on language, as this has been the domain in which compositionality has traditionally played a central explanatory role (see section \ref{sec:frege_to_fodor}). 
However, there has also been a recent surge of interest in compositionality in vision \citep{LeporiSerrePavlick, ZhouFeinmanLake24}, multi-modal models \citep{CastroZiaiSalujaEtAl24, HsiehZhangMaEtAl23, LewisNayakYuEtAl23b, MaHongGulEtAl23, MaLiLiang23, MitraHuangDarrellEtAl23, OssowskiJiangHu24, ShukorRameDancetteEtAl23}, and in reinforcement learning \citep{BakirtzisSavvasTopcu22, JothimuruganBansalBastaniEtAl21, LiuFrank22, ZikelicLechnerVermaEtAl23}, where similar issues have been explored. 
Furthermore, while we have chosen to focus our review on metalearning and LLMs, it should be noted that much recent work has pursued other approaches. 
For example, researchers have investigated whether certain architectural inductive biases can facilitate compositional generalization \citep{AltabaaWebbCohenEtAl23, CsordasIrieSchmidhuber22, KrieteNoelleCohenEtAl13, PalangiSmolenskyHeEtAl17, RougierNoelleBraverEtAl05d, RussinJoOReillyEtAl, SoulosHuMccurdyEtAl23, SchlagSmolenskyFernandezEtAl19, WebbFranklandAltabaaEtAl24, WebbHolyoakLu23}, explored techniques involving data augmentation \citep{Andreas20, CazzaroLocatelliQuattoni24a, ChaiLiLiuEtAl23, JiangZhouBansal22} and the injection of structural knowledge via additional supervision or auxiliary tasks \citep{ChakravarthyRussinOReilly22, JiangBansal21}, and investigated the limits of standard transformers on compositional tasks \citep{DziriLuSclarEtAl23, ZhouBradleyLittwinEtAl23}.

After the review, we discuss how these new approaches may inform our understanding of human compositionality, including their implications for its neural mechanisms and development. 
We conclude with a discussion of the `mere implementation' objection, the second horn of Fodor and Pylyshyn’s dilemma.

\subsection{Metalearning}
\label{sec:metalearning}
In metalearning \citep{BinzDasguptaJagadishEtAl23, GriffithsCallawayChangEtAl19, SandbrinkSummerfield24, WangKurth-NelsonKumaranEtAl18, Wang20}, a model is trained not on a single task but on a distribution of tasks, giving it the opportunity to \emph{learn how to learn} new tasks more rapidly and to leverage regularities among tasks to generalize in specific ways. 
Metalearning can impart specific inductive biases to a model.
For example, a metalearning agent trained on a distribution of navigation tasks may learn exploration strategies that are specifically effective for that distribution, thus allowing it to learn more efficiently how to navigate in similar environments.

Importantly, these inductive biases are imparted to the model without specifying any particular algorithmic features in advance \citep{McCoyGrantSmolenskyEtAl20}. 
Instead, the researcher can simply define the set of learning problems at which the model should excel and rely on the optimization process during metalearning to discover whatever inductive biases are useful for accomplishing these behaviors. 

A number of metalearning methods have been developed for use with neural networks \citep{BengioBengioCloutier91, BengioBengioCloutier95, DuanSchulmanChenEtAl16, FinnAbbeelLevine17, HospedalesAntoniouMicaelliEtAl20, Schmidhuber87}. 
In many of these methods, the model learns to implement a separate ``inner-loop'' or ``in-context'' learning (ICL) algorithm in its forward activation dynamics, distinct from the ``outer-loop'' or ``in-weight'' learning (IWL) algorithm used to train the network in the first place.
In this case, the network’s in-context learning algorithm ``learns'' within the flow of information from its inputs to its outputs \citep{DuanSchulmanChenEtAl16, Lake19, LakeBaroni23, SantoroBartunovBotvinickEtAl16, WangKurth-NelsonKumaranEtAl18}. 
The network’s weights are trained such that the forward pass of the model can take some study examples as inputs in context and learn from them to produce the correct generalization behavior on a test query.
For example, a sequence-processing network such as an RNN or a transformer can be trained on inputs that consist of (1) a set of study examples along with their labels, and (2) a single test query whose label is not given, but must be inferred. 
In this case, the network is trained to produce a prediction about the label corresponding to the test query, conditioned on the set of training samples and their labels. 
The model is then evaluated by freezing its weights and testing on new study examples and a new test query drawn from another task. 

In this setting, we can distinguish the IWL algorithm that determines the updates to the network’s weights during metalearning, and the ICL algorithm that is implicitly implemented in the network’s activation dynamics.
The ICL algorithm does not require updates to the weights of the network to learn (i.e., it can still learn a new task when the weights are frozen). 
Crucially, these two distinct learning algorithms can have different properties.
For example, the ICL algorithm may have a completely different updating procedure or effective learning rate, or may embody inductive biases that are not present in the original IWL algorithm. 

This suggests a novel perspective on how a neural network might instantiate compositionality without incorporating explicit classical symbols: by (meta)learning an ICL algorithm with an inductive bias for compositional generalization. 
Even if the usual IWL algorithm used by neural networks lacks compositionality, as theorists such as \citet{FodorPylyshyn88} and \citet{Marcus98} suggest, it is still possible that the ICL algorithm that a network comes to implement in its activation dynamics will exhibit compositionality. 

\citet{LakeBaroni23}, following up on earlier work \citep{Lake19}, applied this approach to compositional generalization problems such as the SCAN task \citep{LakeBaroni18}, and showed that a metalearning transformer was capable of near-perfect accuracy on the task. 
The transformer was trained on a distribution of tasks, each a compositional generalization problem in its own right.
In particular, each task had a few samples from the grammar that were in principle sufficient to infer the rules and produce the correct answer on the test sample. 
The model was trained on some generalization tasks, and tested on ones that it had not seen. 
The results showed that the transformer was capable of metalearning to implement a compositional learning algorithm in its activation dynamics, achieving near-perfect accuracy on held-out compositional generalization tasks that were not seen during metalearning.

These results demonstrate a novel strategy for endowing neural network models with the property of compositionality, and for overcoming the problem of o.o.d. generalization. 
After metalearning was complete, the model was given an o.o.d. problem in context, and its ICL (inner-loop) algorithm exhibited an ability to generalize. 
However, from the perspective of the IWL (outer-loop) algorithm, the problem was familiar (i.i.d.), because it was trained on similar o.o.d. generalization tasks during metalearning.
Thus, metalearning creates a scenario where two kinds of generalization can coexist simultaneously: the IWL algorithm is simply generalizing i.i.d., while the ICL algorithm is capable of generalizing o.o.d.
This o.o.d. generalization is possible because of the inductive biases imparted to the ICL algorithm throughout metalearning.

\citet{LakeBaroni23} conclude that although their metalearner could generalize on o.o.d. problems that were similar to those seen during metalearning, it could not generalize to completely new kinds of problems (i.e., ones that were o.o.d. with respect to the metalearning dataset). 
One way of interpreting this finding is to maintain that the kind of compositionality that humans exhibit is truly o.o.d., so the model’s failure shows that it is inadequate as a model of human compositionality. 
However, another interpretation invites a reexamination of human compositionality and suggests that it too could be a property of an ICL algorithm, in this case implemented by the human brain \citep{RussinMcGrathPavlickFrank24}. 
This would challenge the claim that humans are capable of generalizing compositionally in completely novel domains, instead establishing the weaker claim that humans generalize o.o.d., but only in specific domains determined either by evolutionary history or by prior learning experiences.

\subsection{Large Language Models}
\label{sec:llms}
The original tasks used to evaluate compositionality \citep[e.g. the SCAN dataset;][]{LakeBaroni18} were developed at a time when most models were trained from scratch to perform a specific task, such as machine translation \citep{BahdanauChoBengio14, VaswaniShazeerParmarEtAl17a}. 
However, researchers soon began to utilize self-supervised methods that did not require human labels (e.g., annotations, translations, etc.), thus allowing models to be trained on the huge amounts of unlabeled text available on the internet. 
The dominant self-supervised task has been language modeling, where a neural network parameterizes a distribution over sequences of natural language text. 
For example, in causal language modeling this distribution is expressed in terms of the probability of the next token given the entire preceding context. 
Researchers found that as transformer models were scaled up (in terms of number of parameters and amount of data) language modeling performance continued to improve. 
This meant that better performance could be achieved if LLMs were pretrained on unlabeled text and then fine-tuned on the specific tasks of interest, rather than training from scratch on the specific task \citep{DevlinChangLeeEtAl19, PetersNeumannIyyerEtAl18, RadfordWuChildEtAl}. 

Researchers then discovered that at even larger scales (e.g., at the scale of GPT-3, 175 billion parameters), you could in some cases forgo the fine-tuning step and simply ask these LLMs to do a new task \textit{in context} by providing a few examples of the task to be performed \citep{BrownMannRyderEtAl20a}. 
For example, rather than fine-tuning a pretrained model to predict whether a human-labeled restaurant review was positive or negative, you could prompt the model with a few study examples consisting of reviews and their corresponding labels, and evaluate it on a test query---an unlabeled review.
These emergent ICL abilities allow LLMs to flexibly solve new tasks that are explicitly instructed or demonstrated with a few examples, and is further improved when models are fine-tuned on human-generated instruction-following datasets \citep{WeiBosmaZhaoEtAl22} and with reinforcement learning by human feedback 
\citep[RLHF;][]{OuyangWuJiangEtAl22}. 
This created yet another paradigm shift in natural language processing: very large models are trained on very large datasets and then tested in context on specific tasks. 

\begin{figure}[t]
    \centering
    \includegraphics[width=0.8\textwidth]{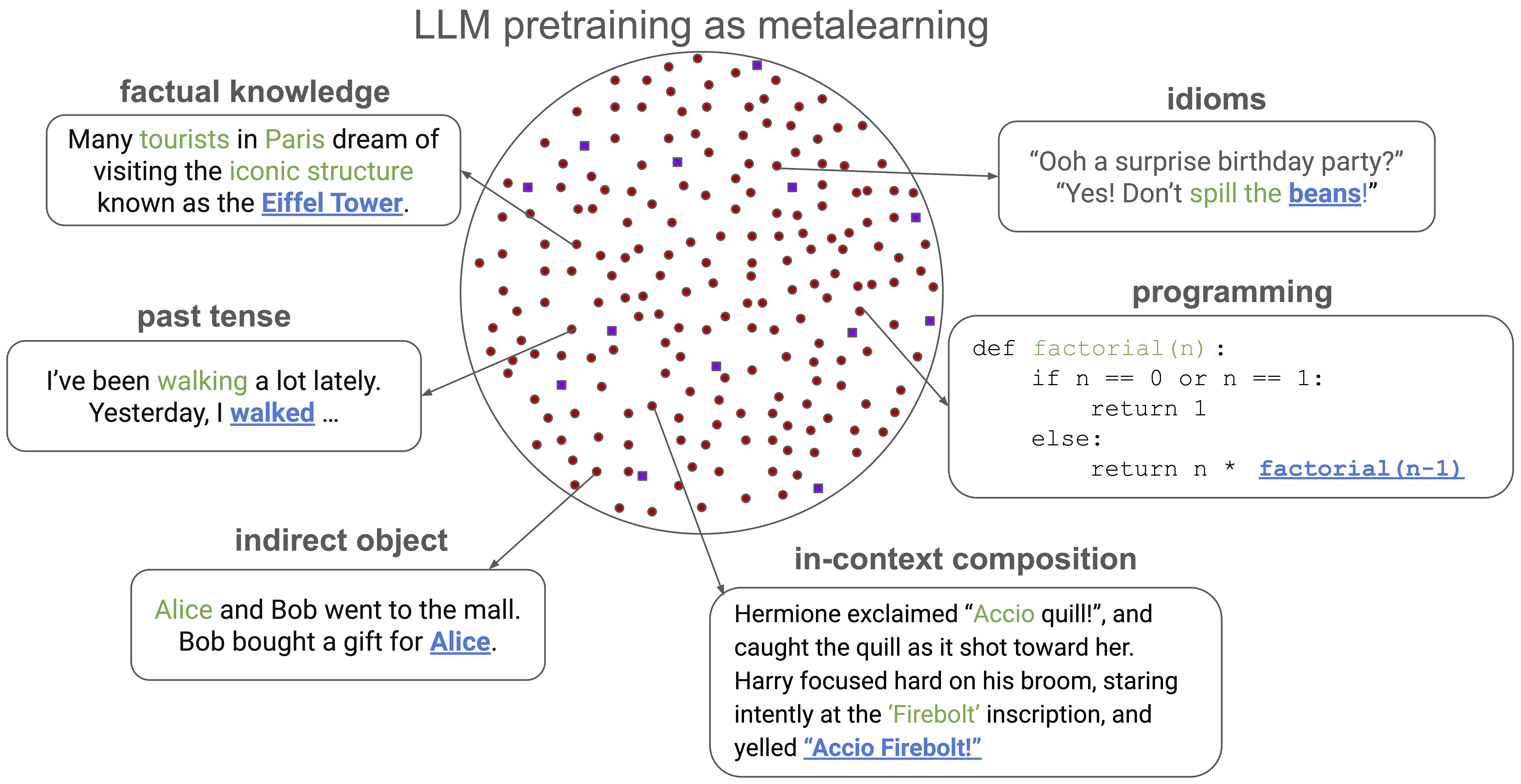}
    \caption{Predicting the next word on a large corpus of text can be seen as a kind of metalearning \citep{{BrownMannRyderEtAl20a, RadfordWuChildEtAl}}, in which the LLM trains on tasks requiring diverse capabilities such as memorization of factual knowledge, idioms, inflectional morphology (past tense), programming, and identifying the indirect object of a sentence \citep{WangVariengienConmyEtAl22}. In some cases, successful prediction of the next word may require \emph{in-context composition} of novel elements. As in metalearning, a distinction can be made between the IWL algorithm training the weights of the network, and an implicit ICL algorithm that operates on information given in context to make inferences relevant to predicting the next word(s).}
    \label{fig:llm_distribution}
\end{figure}

From a theoretical perspective, this “pretrain-test” paradigm \citep{HupkesGiulianelliDankersEtAl23} has important similarities with the metalearning setting described above. 
In metalearning, models acquire inductive biases by training on a distribution of tasks. 
The pretraining step in LLMs, where they are trained to predict the next word on a very large dataset, can be seen as training on a distribution of tasks \citep[see Figure \ref{fig:llm_distribution};][]{BrownMannRyderEtAl20a, ChanSantoroLampinenEtAl22, RadfordWuChildEtAl}.
Sometimes this prediction will look more like guessing based on frequencies (e.g., memorizing a common idiom), but other times the task can involve more sophisticated behaviors such as understanding recursion in computer programming.
There are doubtless many instances in these large datasets where the best way to predict the next word is to learn something in context, and perhaps even to compose novel concepts in context (see “in-context composition” in Figure \ref{fig:llm_distribution}). 
Thus, although it is easy to dismiss next-word prediction as a simplistic or purely statistical training objective, upon further reflection one can see how it might put pressure on the model to develop an ICL algorithm in its forward activation dynamics, and perhaps even to endow this ICL algorithm with the property of compositionality.

LLMs at this scale perform quite well on many kinds of tasks that seem to require compositionality. 
As anyone who has interacted with chatGPT or similar products knows, these models are capable of writing whole paragraphs of coherent text about almost any topic, including novel concepts or made-up words that are included in the prompt. 
They can construct sentences with sophisticated grammatical structure \citep{LinzenBaroni21a}, and can put together well-reasoned arguments \citep{HerboldHautli-JaniszHeuerEtAl23}. 
They even recapitulate some of the syntactic phenomena thought to require innate grammatical constraints such as syntactic island effects \citep[][ although see \citealt{LanChemlaKatzir24}]{WilcoxFutrellLevy23}. 
They are also adept at computer programming, and are capable of writing novel functions or scripts (e.g., involving compositions of known functions) to accomplish a goal given in natural language instructions. 
These models have achieved state-of-the-art performance on natural language inference tasks such as recognizing logical entailment \citep{BrownMannRyderEtAl20a}, and have even demonstrated human-like tendencies in analogical reasoning tasks such as Raven’s Progressive Matrices \citep{WebbHolyoakLu23}, which arguably require compositionality as well. 

Researchers have also evaluated these models on the tasks designed specifically to test compositional generalization capabilities \citep{LakeBaroni18}.
As in the metalearning setting, the evaluation works slightly differently than in the original SCAN dataset, where models were trained from scratch on tens of thousands of examples without ``jump'' and tested on instructions containing ``jump.'' 
In the pretrain-test setting, the pretrained LLMs are given just a few study examples from the SCAN task in context and asked to generalize to new test samples that are also given in the same input. 
The compositional generalization test can be replicated by excluding certain kinds of examples from the context (e.g., longer examples, or more complicated constructions involving ``jump''). 
When GPT-3 (code-davinci-002) was evaluated with a standard prompt on a compositional split of the SCAN task in this way, it only achieved 16.7\% accuracy \citep{ZhouScharliHouEtAl23}.
However, the inclusion of novel prompting methods improved performance to a near-perfect accuracy of 99.7\%. 
These prompting methods, called ``chain-of-thought'' prompting \citep{WeiWangSchuurmansEtAl23} and ``least-to-most'' prompting \citep{ZhouScharliHouEtAl23}, have been shown to improve performance on many reasoning tasks. 
In chain-of-thought prompting \citep{KojimaGuReidEtAl23, WeiWangSchuurmansEtAl23}, the model is asked to ``think step-by-step,'' and/or given examples that show intermediate steps of reasoning. 
In least-to-most prompting \citep{DrozdovScharliAkyurekEtAl22, ZhouScharliHouEtAl23}, these in-context examples are simply given in order of least difficult to most difficult.
Many other prompting strategies have been developed to further improve performance on similar tasks \citep{HuangChang23, NyeAndreassenGur-AriEtAl21, PressZhangMinEtAl23, YaoYuZhaoEtAl23}. 

The fact that the prompting strategy makes such a difference in the resulting measures of performance may be taken to indicate that the compositional generalization exhibited with certain prompts is fragile and cannot reflect a genuine competence \citep{BenderKoller20, BenderGebruMcMillan-MajorEtAl21, LanhamChenRadhakrishnanEtAl23, TurpinMichaelPerezEtAl23, WebsonPavlick22}.
However, it is well understood that human performance depends in large part on the instructions they are given \citep{EvansNewsteadAllenEtAl94}, and the specific content or domain on which their abilities are tested \citep{KlauerMuschNaumer00, Wason68}. 
LLMs exhibit similar content effects \citep{DasguptaLampinenChanEtAl23}, highlighting the need for fair comparisons between humans and machines \citep{Firestone20, Buckner23}.

Aside from these language-specific compositional generalization settings, much work has focused on evaluating whether large vision models or vision-language models exhibit similar compositional abilities \citep{MaHongGulEtAl23, ConwellUllman22, ZhengZhangKembhaviEtAl24}.
The results here have been mixed, suggesting a dissociation between the performance measured on compositional generalization problems developed in language-specific settings and those developed in other domains.

Taken together, the impressive performance of LLMs shows that at least in some cases, the behavioral signatures of compositionality can emerge from the ICL algorithm implemented in the forward activation dynamics of a large model trained to predict the next word on a large corpus of text.
Similar to the studies using metalearning models \citep{Lake19, LakeBaroni23}, these LLMs demonstrate that even though the IWL algorithm may not itself be capable of compositional (o.o.d.) generalization, it can endow an emergent ICL algorithm with this capability. 
This again shows that the o.o.d. generalization problem solved by the ICL algorithm may be an i.i.d. generalization problem from the perspective of the original IWL algorithm, which has probably had many opportunities to learn how to learn similar problems in context from its training set (see Figure \ref{fig:llm_distribution}). 

Can these findings about LLMs inform us about human compositionality? 
It is of course difficult to make any inferences from LLMs to humans because there are so many differences between them. 
For example, transformers are not biologically plausible in many ways, and the scale of LLM training datasets is orders of magnitude larger than a human lifetime’s worth of language data \citep{Frank23, LinzenBaroni21a, Pavlick23, WarstadtBowman24}. 
However, these results do show that in principle it is possible to metalearn an ICL algorithm capable of compositional generalization by simple self-supervised learning on natural language. 
Furthermore, in contrast with explicit metalearning \citep{LakeBaroni23}, the impressive compositional behaviors of LLMs show that the distribution of tasks on which a metalearner is trained does not need to be deliberately contrived to encourage compositionality---strong compositional generalization abilities emerge in these models even when they are trained on unstructured text. 
This suggests that humans, who also learn from relatively unstructured experiences, could plausibly metalearn throughout development to implement a compositional ICL algorithm as well \citep{RussinMcGrathPavlickFrank24}.

\section{Discussion}
\label{sec:discussion}

Classical cognitive theorists like \citet{Chomsky65} and \citet{Fodor75} developed compelling arguments that compositionality is a central property of human cognition, crucial for explaining the creativity, productivity, and systematicity of language and thought (see section \ref{sec:frege_to_fodor}). 
Neural network modeling was dismissed as an empirically inadequate explanatory paradigm because these models seem to lack the kind of combinatorial constituent structure intrinsic to classical architectures based on atomic symbols and governed by syntactic rules \citep[see section \ref{sec:FandP};][]{FodorPylyshyn88}. 

Even as deep neural networks blew past competing frameworks in virtually every area of AI, initial investigations into their compositionality seemed to confirm traditional intuitions that they were not capable of replicating the kinds of compositional generalization that humans exhibit \citep[see section \ref{sec:operationalizing};][]{LakeBaroni18, KimLinzen20a}. 
One way of understanding these initial findings is from the statistical learning perspective: compositional generalization requires out-of-distribution (o.o.d.) generalization or extrapolation, but neural networks assume that train and test data will be independent and identically distributed (i.i.d.). 
Standard neural networks fail on these problems, then, because they lack the strong inductive biases that would allow them to generalize outside of their training distributions. 
Some have taken these results to demonstrate that neural networks will never achieve compositionality on their own and concluded that they must therefore be augmented with classical symbolic processes \citep[``neuro-symbolic hybrids;''][]{MarcusDavis20, Marcus20b}. 

More recent advances complicate this picture, suggesting alternative ways in which strictly neural models might explain the same generalization behaviors \citep{Baroni20}. 
In particular, recent results suggest that metalearning (section \ref{sec:metalearning}) and large-scale pretraining in LLMs (section \ref{sec:llms}) can endow neural network models with sophisticated in-context learning abilities, and that these abilities can in some cases capture compositional generalization behaviors.

It can be argued that none of these neural networks are actually generalizing outside of their training distributions, and that they have therefore failed to explain human generalization capabilities. 
However, another perspective on these results is that they show that the generalization behaviors taken as diagnostic of compositionality in humans are possible without strongly out-of-domain generalization.

More generally, these findings provide a novel perspective on human compositionality, suggesting that metalearning and in-context learning may be important principles for explaining key aspects of these generalization behaviors in humans. 
In the following, we briefly discuss how these principles can provide novel empirical predictions about the neural mechanisms and development of human compositionality. 

\subsection{Metalearning and Neural Mechanisms}
\label{sec:neuro}

The metalearning approach hypothesizes that human compositionality is a property of an in-context learning (ICL) algorithm that has been metalearned via the usual in-weight learning (IWL) \citep{BinzDasguptaJagadishEtAl23, DubeyGrantLuoEtAl20, LakeBaroni23, RussinMcGrathPavlickFrank24, WangKurth-NelsonKumaranEtAl18}.\footnote{Some have assumed that metalearning in neural networks is a better model of learning on an evolutionary timescale than within an individual’s lifetime \citep{BinzDasguptaJagadishEtAl23}. Here, we assume that metalearning can take place in the brain over the course of an individual’s lifetime, and that ICL and IWL can simultaneously exist within a single system. This idea is directly investigated by \citet{RussinPavlickFrank24}.}
This hypothesis predicts that when the neural mechanisms supporting ICL, rather than IWL, are driving behavior, this should facilitate compositional generalization. 

\citet{RussinPavlickFrank24} found that this interplay between ICL and IWL in neural network models explained compositional generalization performance on a task recently studied in humans \citep{DekkerOttoSummerfield22c}. 
Both LLMs and metalearning neural networks achieved much better compositional generalization performance on the task when ICL rather than IWL was responsible for behavior. 
Furthermore, \citet{RussinPavlickFrank24} found that the dynamic interaction between ICL and IWL \citep{ChanSantoroLampinenEtAl22}, which has similarities to a tradeoff between working memory and reinforcement learning observed in humans \citep{CollinsFrank18a, Rac-LubashevskyCremerCollinsEtAl23}, also explained the curriculum effects observed in the human study \citep{DekkerOttoSummerfield22c}.  

There is further support from cognitive and computational neuroscience for the idea that ICL and compositional generalization share a set of underlying neural mechanisms in humans. 
The prefrontal cortex is important for basic capacities known to be involved in human ICL, such as working memory \citep{OReillyFrank06}, top-down attention \citep{CohenDunbarMcClelland90}, and cognitive control \citep{MillerCohen01}, and, as previously mentioned, is thought to be important for compositional generalization processes such as rule learning \citep{CalderonVergutsFrank22, CollinsFrank13, FrankBadre12, KrieteNoelleCohenEtAl13}, reasoning \citep{CrescentiniSeyed-AllaeiPisapiaEtAl11, Goel07}, and processing complex syntax \citep{Thompson-Schill05}. 
Furthermore, it has been hypothesized that metalearning (specifically, meta-reinforcement learning) is an important aspect of the functioning of the prefrontal cortex \citep{HattoriHedrickJainEtAl23, WangKurth-NelsonKumaranEtAl18}.

Thus, while it may be difficult to directly establish that human compositional generalization behaviors are emergent properties of an ICL algorithm, there is indirect empirical support for the idea that compositional generalization and ICL share an underlying set of neural mechanisms in humans. 

\subsection{Metalearning and Human Development}
\label{sec:development}

The metalearning approach emphasizes how inductive biases can themselves be learned from data. 
As has been articulated clearly in other work \citep{BinzDasguptaJagadishEtAl23, LakeBaroni23, McCoyGrantSmolenskyEtAl20, Wang20}, the metalearning approach is agnostic about whether this occurs on an evolutionary timescale or within an individual human lifetime. 
Many have shied away from interpreting metalearning neural networks as models of human development \citep[although see ][]{RussinMcGrathPavlickFrank24, Wang20, WangKurth-NelsonKumaranEtAl18}, instead choosing to emphasize its utility as a methodological tool \citep{BinzDasguptaJagadishEtAl23, McCoyGrantSmolenskyEtAl20}.

When taken seriously as models of human development, however, metalearning models generate testable empirical predictions. 
If human compositionality is the result of metalearning, we should expect older children to show signs of improved compositional generalization abilities. 
Existing evidence supports this prediction, suggesting that children learn how to learn more efficiently over time \citep{Bergelson20}, and that the specific ability to compose rules improves over the course of development \citep{PiantadosiAslin16, PiantadosiPalmeriAslin18}.

A more specific prediction of the metalearning hypothesis is that ICL, as opposed to IWL, should improve with experience. 
This is also consistent with existing evidence, which suggests that older children are much more adept at learning and reasoning in context using working memory and other executive functions \citep{MunakataSnyderChatham12}. 
Indeed, executive functions, which are intimately related to our ability to learn in context \citep{ColeEtzelZacksEtAl11, DuncanChylinskiMitchellEtAl17, HampshireThompsonDuncanEtAl11a, MillerCohen01, RougierNoelleBraverEtAl05d}, emerge especially late in development and continue to mature throughout adolescence and early adulthood \citep{FergusonBrunsdonBradford21}.

Finally, the metalearning hypothesis emphasizes that compositional generalization abilities are facilitated by exposure to similar generalization problems. 
This predicts that humans will be more likely to exhibit compositional generalization behaviors with concepts or stimuli that they have previously encountered. 
The extensive literature on category learning has demonstrated that humans generalize better when categories are congruent (or verbalizable) with familiar features \citep{AshbyMaddox11, FleschBalaguerDekkerEtAl18a, FleschJuechemsDumbalskaEtAl22, LoveMedinGureckis04}. 
It has also been shown that in domains such as motor learning, humans reliably fail to generalize outside of their training distributions on certain tasks \citep{ZhouFitzgeraldColucci-ChangEtAl17}.

These results are consistent with the idea that humans are capable of generalizing out-of-distribution in certain domains where they have been trained to do so (e.g., learning certain kinds of new words or categories), but fail to generalize in completely novel scenarios. 
The question of whether a given task is truly out-of-distribution with respect to a lifetime of human experience is difficult to establish empirically. 
Better naturalistic dataset collection, including egocentric video and audio recordings from children \citep{SullivanMeiPerforsEtAl21, VongWangOrhanEtAl24}, and efforts to curate more ecologically/psychologically plausible language datasets \citep{WarstadtBowman24, WarstadtChoshenMuellerEtAl23, WarstadtMuellerChoshenEtAl23}, should facilitate this line of inquiry.

It is important to note that humans learn from experiences that are highly unstructured compared to the training datasets used in most studies on metalearning \citep[e.g.,][]{Lake19, LakeBaroni23, RussinPavlickFrank24}, which are contrived by researchers to engender very specific inductive biases on very specific tasks. 
While humans are trained on structured curricula in formal educational settings, there is evidence that competences related to compositionality such as sensitivity to hierarchical structure in syntax \citep{Chomsky57} and to geometric features in visual tasks \citep{DehaeneAlRoumiLakretzEtAl22} do not depend on such experiences. 
However, LLMs have shown that simply predicting the next word on a large dataset of unstructured text can engender an inductive bias for compositionality \citep{BrownMannRyderEtAl20a, BubeckChandrasekaranEldanEtAl23, ZhouScharliHouEtAl23}. 
These models develop ICL abilities \citep{vonOswaldNiklassonSchlegelEtAl23a, XieRaghunathanLiangEtAl22} that allow good generalization performance on compositional tasks \citep{WebbHolyoakLu23, WeiWangSchuurmansEtAl23, ZhouScharliHouEtAl23}, even though their training sets were not specifically designed to achieve this. 
These results are impressive, and provide a proof of concept that an inductive bias for compositionality can be learned from unstructured language data.

However, current LLMs are trained on orders of magnitude more data than humans experience in an entire lifetime \citep{Frank23, LinzenBaroni21a, WarstadtBowman24}, making it unclear whether similar capabilities could emerge in models trained on a more realistic scale. 
The emergence of ICL abilities has been shown to depend on the distributional properties of the training data and on network architecture \citep{ChanSantoroLampinenEtAl22}.
Furthermore, architectural inductive biases such as those related to the ``relational bottleneck'' \citep{WebbFranklandAltabaaEtAl24} have been shown to improve sample efficiency on certain tasks \citep{WebbSinhaCohen21}. 
Thus, perhaps human compositionality is best characterized as an emergent property of an in-context learning algorithm, but the brain embodies important architectural inductive biases that encourage this algorithm to emerge.

\subsection{Mere Implementations?}
\label{sec:FandP2}
The empirical evidence reviewed up to this point suggests that neural network models may be able to get past the first horn of Fodor and Pylyshyn’s critical dilemma (see section \ref{sec:FandP}). 
We have focused on this initial empirical challenge because we believe that recent advances represent a crucial development on this front, eliminating a major obstacle that has hindered the acceptance of neural networks as viable models of human cognition. 
But what about the `mere implementation' objection, the second horn of the compositionality challenge? 
Is the success of contemporary DNNs on compositional generalization problems a reflection of the fact that they have implemented classical architectures? 
And if so, does this mean that they are uninformative for cognitive theory?

Though this is a nuanced and challenging issue, certain cases are unambiguous. 
Neuro-symbolic hybrid models \citep{EllisWongNyeEtAl21, GaurGunaratnaBhattEtAl22, Marcus20b, MarcusDavis20, YuYangLiuEtAl23, ZhouFeinmanLake24}, which combine explicit symbolic processes with neural networks, are `mere implementations' by design---proponents of this approach treat the behavioral signatures of compositionality in humans as diagnostic of underlying symbolic mechanisms and explicitly build such mechanisms into neural networks. 
Though the neural components of these networks are utilized to deal with high-dimensional inputs like images, their compositionality is entirely due to their built-in symbolic components.
This captures something of the spirit of Fodor and Pylyshyn’s original objection (in fact, they allude to this possibility themselves), and transforms it into an empirical hypothesis with clear predictive upshot---explicit implementation of symbolic mechanisms will be a necessary prerequisite for neural networks to succeed on compositional problems. 
Current results do not seem to favor this hypothesis, as the findings on metalearning (section \ref{sec:metalearning}), and LLMs (section \ref{sec:llms}) suggest that such explicit symbolic mechanisms may not be necessary. 
However, it is still unclear whether neuro-symbolic hybrids will win out in the end. 
If so, it would mean that DNNs yield few novel insights into compositionality, though they may still offer alternative perspectives on other aspects of cognition \citep{Chalmers93}.

However, the claim that neural networks may implement classical symbolic operations is invoked to stave off challenges from DNNs, even in the absence of any tangible predictions of this kind. 
Without clear mechanistic commitments, it becomes harder to assess. 
The results reviewed here indicate that even strictly neural models with no built-in symbolic components may be adequate for replicating the behavioral signatures of compositionality. 
What should we make of this outcome? 

One possibility that was widely discussed in the original connectionist debate is eliminativism \citep{RamseyStichGaron90}. 
An eliminative connectionist would argue that the success of strictly neural models demonstrates that Chomsky and Montague's shift to treating formal syntactic and semantic theories as models of human psychological processing (see section \ref{sec:comp_in_context}) was a grave mistake \citep{Piantadosi23}, and that the ideal language philosophers' skepticism about this approach was well-founded. 
If DNNs can achieve compositional generalization without internally representing any rule-based, combinatorial systems, this undermines the motivation for positing them to explain human behavior. 
Chomsky was right that this capacity cries out for explanation, but the eliminativist holds that the neural network vocabulary of nodes, weights, layers, and so forth, will furnish sufficient explanatory resources for doing so.

The possibility that these networks are implementing classical architectures serves as an obstacle to such eliminativist conclusions \citep{McGrathRussinPavlickEtAl23, McGrathRussin2024}. 
Physical systems can exhibit functional organization at different levels of description. 
The fact that no explicit symbols or syntactic operations were built into a DNN that has metalearned to generalize compositionally doesn’t mean that it hasn’t learned to implicitly implement these classical constructs through training. 
For all we know---runs this version of the objection---the DNNs that succeed on these compositional tasks are not alternative cognitive models at all, but mere implementations of the very same theoretical constructs that classical theorists have emphasized all along. 
On this view, they may help us to understand how classical architectures are implemented in the human brain \citep{McLaughlin93}, but will not be informative at the cognitive level.

Though this is an important empirical possibility, there is a crucial gap between the claim that a pretrained LLM \emph{could, for all we know,} be merely implementing a classical symbolic architecture and any claim that it \emph{must be}. 
The problem with Fodor and Pylyshyn’s dilemma is that it treats mere implementation as a virtual guarantee---if neural networks ever overcome the first horn and exhibit the behavioral signatures of compositionality, it must be because they have merely implemented classical architectures (thereby falling on the second horn). 
There may be arguments that narrow this gap,\footnote{One way to bridge the gap between these two claims would be to argue that the “possibility space” \citep{CaoYamins21} is narrow, so that DNNs are likely to be hitting on the very same solutions postulated by classical theorists. One issue with this approach is how to justify the claim that the space of possible solutions is narrow---reliance on intuition here may reflect a failure of contemporary imagination, rather than any genuine limitation. 
Moreover, even if the solution set is limited, it need not be limited to a single possibility, leaving room for interesting and informative variation between the solutions devised by researchers and those identified by the models themselves.} but Fodor and Pylyshyn’s blithe confidence is unwarranted.

The only way to \emph{guarantee} that the success of DNNs reflects mere implementation would be to treat the behavioral signatures of compositionality as constitutive (rather than diagnostic) of an underlying classical architecture \citep{McGrathRussinPavlickEtAl23a}. 
\emph{What it is} to be a classical architecture, on this constitutive view, \emph{just is} to be the kind of system that generates this behavioral profile. 
Adopting this position would ensure that any successful DNN would count as a mere implementation of a classical architecture, but it would do so at the cost of sapping classical theory of the kind of explanatory value that it was originally intended to contribute. 
The rise of cognitivism and its attendant rejection of behaviorism hinge on the claim that psychological theory must do more than merely redescribe behavior---we need an understanding of the independently identifiable underlying mechanisms that produce the given behavior \citep{CaoYamins21b}. 
If giving rise to these generalization behaviors is constitutive of being a classical architecture, however, then we preclude even the logical possibility that a DNN could replicate the behavioral signatures of compositionality without implementing a classical system. 
This means, in effect, that we have failed to specify any theoretical content beyond this characterization of behavior and have lapsed back into mere redescription.

Rather than simply assuming that contemporary DNNs implement classical architectures or searching for some \emph{a priori} guarantee, what we need to do is try to \emph{find out} whether or not they are doing so. 
The burgeoning field of ``mechanistic interpretability'' is making this increasingly feasible \citep{elhage2021mathematical, OlssonElhageNandaEtAl22, MerulloEickhoffPavlick24, WuGeigerIcardEtAl}. 
Though interpretability work is often raised in the context of AI safety considerations \citep{RudnerToner21}, it can also play a crucial role in informing cognitive theory, allowing us to extract high-level, functionally relevant descriptions of these models’ internal representations that we can put into contact with existing cognitive theory \citep{McGrathRussinPavlickEtAl23a, Milliere24, MilliereBuckner24a}. 
In principle, this process could point us towards eliminativism (if successful models’ representations bear no resemblance to classical constructs), or mere implementationism (if they exactly recapitulate the structure of a fully specified symbolic theory). 
In the cases where this work has been pursued, however, it seems likely that these models land somewhere in between, recapitulating certain aspects of classical theory, but diverging in interesting ways \citep{Baroni22, ChenShwartz-ZivChoEtAl24, ManningClarkHewittEtAl20a}. 
In such cases, we have suggested that it may be worth preserving key terms to encompass these behaviors \citep{McGrathRussinPavlickEtAl23, McGrathRussin2024}, as long as it does not obscure the fact that they contain new insights. 
In a slogan, not all systems that we may count as implementations are `mere'---informative implementations can point us to new insights into previously postulated structures \citep{Smolensky88, SmolenskyLegendre11}. 
Though interpretive work evaluating the compositional models reviewed here is still in its initial stages, it is a promising line of future investigation \citep{Andreas19a, LeporiSerrePavlick, LewisNayakYuEtAl23b, McCoyLinzenDunbarEtAl19, RussinFernandezPalangiEtAl21, SoulosMcCoyLinzenEtAl20, ToddLiSharmaEtAl24}.
From a philosophical perspective, further progress on deeper issues about the nature of explanation and implementation will put us in a position to use these findings to resolve the mere implementation objection once and for all.

\subsection{Conclusion}
In this article, we have traced the notion of compositionality from its emergence in early analytic philosophy to its present status as a key explanatory construct in contemporary cognitive science. 
We have emphasized the role that compositionality plays in debates over the explanatory adequacy of neural network models of human cognition and reviewed a range of recent results suggesting that current deep neural networks replicate key behavioral signatures that have motivated the ascription of internal compositional systems to human beings.
In particular, these recent results suggest that metalearning (or large-scale pretraining) is a viable approach to reproducing these behavioral signatures within strictly neural architectures. 
When taken seriously as an approach to modeling human compositionality, the metalearning framework generates concrete predictions about its underlying neural mechanisms and development. 
DNNs' recent success also reopens key philosophical questions about the theoretical significance and proper interpretation of neural networks---whether they should be understood as merely implementing classical compositional mechanisms, or as sources of novel insight that may motivate revisions to classical theories.
These pressing questions take us beyond what neural networks can do, impelling us to ask what they mean for our understanding of the human mind. 
We expect that resolving them will require an interdisciplinary approach and increased philosophical engagement.

\subsection{Acknowledgments}
We would like to thank Lotem Elber-Dorozko, Richard Kimberley Heck, Christopher Hill, Elizabeth Miller, Joshua Schechter, Roman Feiman, Michael J. Frank, Ellie Pavlick, Randall O'Reilly, Edouard Machery, Suraj Anand, Steven Frankland, Taylor Webb, Stavros Orfeas Zormpalas, Felipe De Brigard, Walter Sinnott-Armstrong, and the participants in the Brown University Dissertation Workshop, the Philosophy and Cognitive Science of Deep Learning reading group, and the 2024 SSNAP Workshop for helpful comments and discussions.
JR was supported by NIH NIGMS COBRE grant GR5271961. 

\newpage
\scriptsize
\bibliographystyle{apalike}
\bibliography{references}  
\end{document}